\documentclass[runningheads]{llncs}

\usepackage{eccv}

\usepackage{svg} 
\usepackage{epsfig}
\usepackage{graphicx}
\usepackage{enumitem}
\usepackage{amsmath}
\usepackage{amssymb}
\usepackage{stmaryrd} 
\usepackage{hyperref}

\usepackage[accsupp]{axessibility}  

\def\ie{\emph{i.e.}}
\def\eg{\emph{e.g.}}

\def\etal{\emph{et al.}}
\newcommand{\ours}{RenDetNet}

\begin{document}

\pagestyle{headings}
\mainmatter
\def\ECCVSubNumber{22}  

\title{RenDetNet: Weakly-supervised Shadow Detection with Shadow Caster Verification}
\titlerunning{RenDetNet}

\author{Nikolina Kubiak\inst{1,2} \and
Elliot Wortman\inst{1} \and
Armin Mustafa\inst{1} \and
Graeme Phillipson\inst{2} \and
Stephen Jolly\inst{2} \and
Simon Hadfield\inst{1}}

\authorrunning{Kubiak et al.}

\institute{University of Surrey, Guildford, UK\\
\email{\{n.kubiak,s.hadfield\}@surrey.ac.uk}
\and
BBC R\&D, Salford, UK}

\maketitle

\begin{abstract}
 Existing shadow detection models struggle to differentiate dark image areas from shadows. In this paper, we tackle this issue by verifying that all detected shadows are real, \ie\ they have paired shadow casters. We perform this step in a physically-accurate manner by differentiably re-rendering the scene and observing the changes stemming from carving out estimated shadow casters. Thanks to this approach, the \ours\ proposed in this paper is the first learning-based shadow detection model whose supervisory signals can be computed in a self-supervised manner. The developed system compares favourably against recent models trained on our data. As part of this publication, we release our code on \href{https://github.com/n-kubiak/RenDetNet}{github}.
\end{abstract}


\section{Introduction}
Shadow detection and its twin task of shadow removal are crucial for the development of reliable computer vision systems. Such solutions can be used to solve shadow-related problems in real-life domains such as farming \cite{dornadula2023ai}, remote sensing \cite{ni2022shadow}, autonomous driving \cite{wang_2020_vehicles}, medical diagnostics \cite{yasutomi_2019_shadowmed}, document de-shadowing \cite{jung_2018_documents} or appearance correction in casual capture photos \cite{zhang_portrait_2020}.

\begin{figure}[t]
\begin{center}
\small
\begin{tabular}{  c @{\hspace{0.05cm}} c @{\hspace{0.05cm}} c @{\hspace{0.3cm}}c @{\hspace{0.05cm}} c @{\hspace{0.05cm}} c }
Input& GT SM &  Estim.\ SM & Input& GT SM &  Estim.\ SM\\
\includegraphics[width=1.9cm]{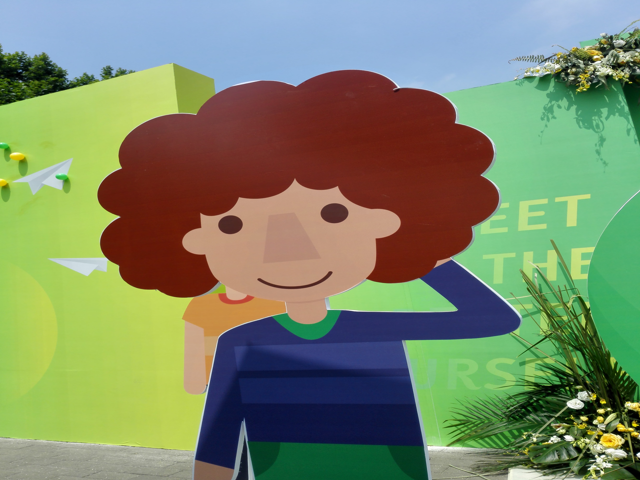}&
\includegraphics[width=1.9cm]{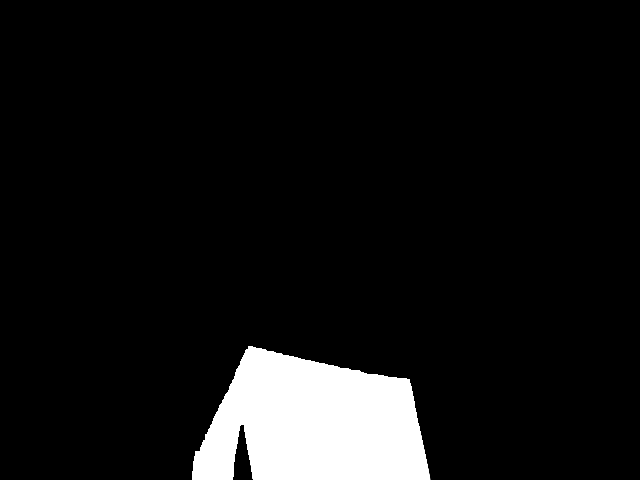}&
\includegraphics[width=1.9cm]{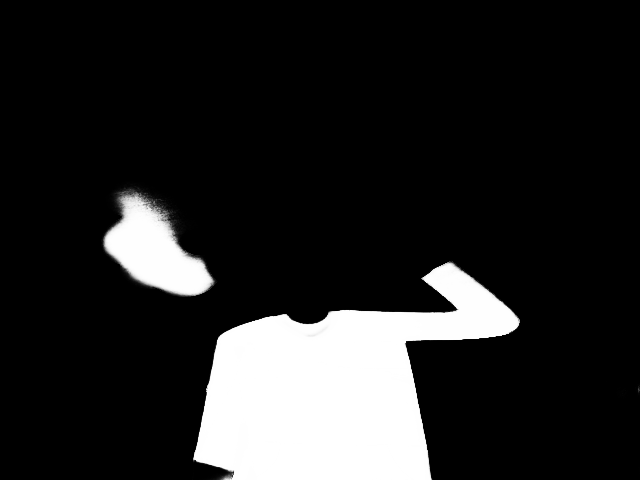}&
\includegraphics[width=1.9cm]{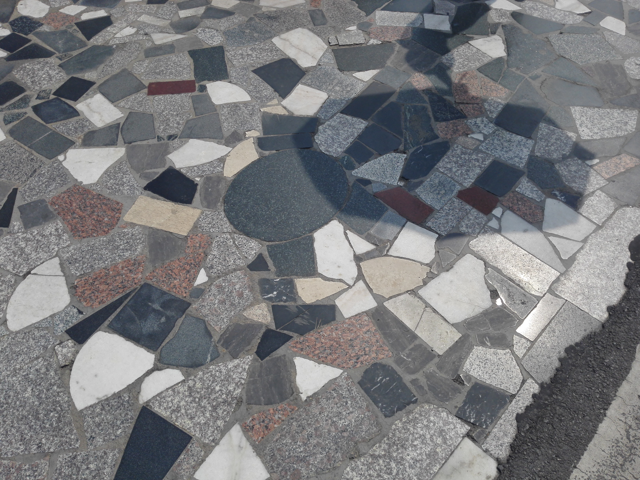}&
\includegraphics[width=1.9cm]{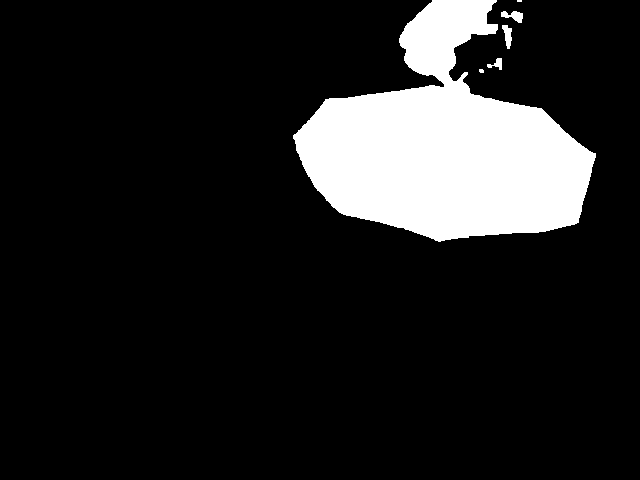}&
\includegraphics[width=1.9cm]{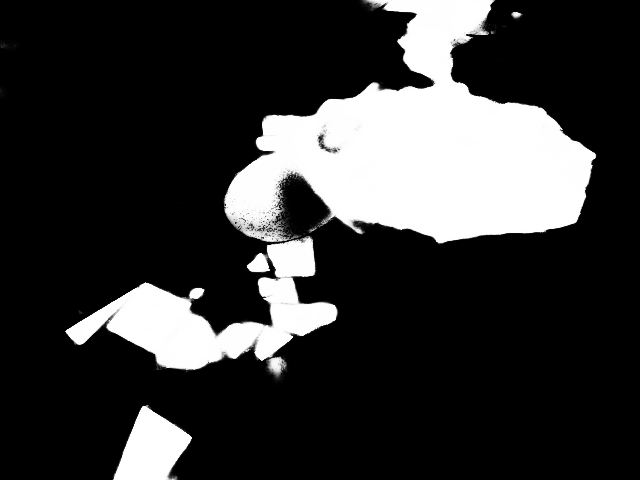}\\
\end{tabular}
\end{center}
\caption{Examples of the common shadow vs dark region problem in shadow detection. The examples come from the ISTD dataset \cite{wang_2018_stacked} and the estimated shadow masks (SMs) were generated using a SOTA shadow detection model \cite{chen2020multi}. This shows that even some of the best available networks still struggle to distinguish between shadows and dark image regions, and inspires us to develop a detection model with improved scene understanding.}
\label{welcome-fig}
\end{figure}

The vast majority of existing shadow detection solutions operate in a supervised manner. The networks take in shadowed images and compare the estimated shadow masks with paired ground truth data. However, this approach has a number of disadvantages: Firstly, the labels are obtained via laborious hand-annotation, and are usually produced by multiple individuals with different labelling behaviours. Secondly, the datasets \cite{vicente2016SBU,wang_2018_stacked,ucf_dataset} that are the \textit{de facto} standard in the image-based shadow detection literature lack diversity. They depict predominantly pavements, grass, etc., and contain little texture variation, other objects and/or clutter. This means that shadow detection is often posed simply as the task of detecting a darker region within the scene. Consequently, when applied to general images, shadow detection models trained on such datasets tend to mistake dark-coloured regions or objects for shadows. This is unsurprising as differentiating shadows from dark regions is an ill-posed problem. Even in biological vision, the only way to solve this problem is through context, and an understanding of scene structure. We visualise this in Fig.\ \ref{welcome-fig}.

In this paper, we propose a completely new way to think about shadow detection. Our model, \ours, is trained without any hand-labelled shadow masks. Instead, our system learns from synthetic rendered scenes, where it estimates a shadow mask and a caster mask (\ie\ a mask of the object casting the shadow). We then use the caster mask to carve out the caster and re-render the scene. If the (binarised) difference between the original render and the carved render matches the estimated shadow and caster regions, it means the masked caster was responsible for the detected shadow. Using this approach, our \ours\ learns not to predict shadows without credible casters.

In summary, the contributions of this paper are as follows:
\begin{enumerate}
    \item We propose a weakly-supervised deep learning approach to shadow detection, producing accurate caster masks and shadow masks corresponding to real shadows; this includes cast and self-cast shadows (Fig.\ \ref{caster})
    \item We present a self-supervised caster-aware dataset generation pipeline;
    \item Our new model compares favourably against recent shadow detection methods on the new shadow-caster datasets presented in this paper.
\end{enumerate}

\section{Related work}
Some shadow detection models treat the task as a preliminary step for shadow removal \cite{wang_2018_stacked,yucel2023lra}. However, more relevant to our work are techniques which attempt to distinguish between dark texture and shadow regions. scGAN \cite{nguyen_2017_scgan} is trained with tunable sensitivity and final masks created via multi-scale mask aggregation. Le \etal\ \cite{le2018a+d} attenuate the strength of shadows in their dataset to train a robust detector. Hu \etal\ \cite{hu2018direction} learn the direction-aware spatial context (DSC) to estimate the lighting direction and thus the regions likely to be shadows. BDRAR \cite{zhu2018bidirectional} refines the features by combining the contexts in adjacent layers, and uses a bidirectional feature pyramid network to aggregate them across the network. Wang \etal\ \cite{wang2018densely} use stacked parallel fusion branches to combine the global context with the fine local features. DSDNet \cite{zheng2019distraction} actively tackles false positives and negatives in shadow detection by operating on multi-scale features. Zhu \etal's model \cite{zhu_2022_complementary} detects shadow and non-shadow areas, and leverages this complementary information to reduce the frequency of misdetections. Jie and Zhang \cite{jie_2022_multi-level} use shuffled multi-level features to provide good global and local context simultaneously. FDRNet \cite{zhu2021mitigating} extracts illumination-invariant and -variant features, and combines them in a way that improves its robustness to brightness changes. 
In FCSD-Net \cite{valanarasu2023fine} the estimated masks are verified through shadow removal. Similarly, SDDNet \cite{cong2023sddnet} models shadows and the backgrounds upon which they fall (\ie\ shadow-free areas) separately. Sun \etal\ \cite{sun2023adaptive} use adaptive mapping to adjust the detection process for raw images. SILT \cite{yang2023silt} tackles the noise in shadow masks using shadow counterfeiting and self-teaching. Wang \etal\ expand the problem of detection, and additionally match all shadow instances with their casters \cite{wang2,wang_2022_inst_tpami,wang1}. The intuition behind these methods is similar to our approach, but the models require full supervision and laborious annotation.

\subsection{Reducing the supervision requirements}
Unlike the previously discussed works, we strive to reduce the necessary supervision requirements to allow for shadow detection in more complex domains. A few other papers explore this idea: A semi-supervised, multi-task MTMT-Net \cite{chen2020multi} learns to estimate shadow regions, edges and count in a supervised way, and is later finetuned using unlabelled data and a consistency loss. Chen \etal\ \cite{chen_2022_semisupvideo} and Lu \etal\ \cite{lu2022video} propose video shadow detection models trained on labelled image data and unlabelled video data; Xing \etal's method \cite{xing2022video} uses labelled and unlabelled videos. Although these approaches use fewer labelled samples, they do not completely eliminate the need for ground truth. We could not identify any deep learning solutions to shadow detection capable of operating in a fully un- or self-supervised manner. However, some non-learning techniques exist, \eg\ that of paired regions. Such methods operate by finding a shadow-free match for each shadowed pixel, based on reflectance consistency \cite{guo_11_paired-regions,wang_21_updated-paired-regions}. The solutions do not require training, but they also do not perform on par with recent deep-learning models. Finally, while the idea of reducing supervision requirements has not been as popular in shadow \textit{detection} works, some examples can be found in the tangential field of shadow \textit{removal} \cite{hu_mask-shadowgan_2019,jin2021dc,kubiak_2024_s3rnet,liu_shadow_2021,Vasluianu_2021_CVPR}.

\subsection{Summary}
As illustrated above, there exists a wide range of shadow detection models. Unfortunately, the supervision requirements of the available solutions are high - the methods are at least semi-supervised. This necessitates a heavy supervision workload for general applicability or, more realistically, leads to systems which can only operate in a restricted domain. Therefore, in this paper we present a novel weakly-supervised model where the necessary supervisory signal can be obtained in a self-supervised manner and the training is guided by differentiable scene carving and re-rendering in synthetic scenes.

\section{Methodology}
\begin{figure*}[t]
    \centering
\includegraphics[width=12cm]{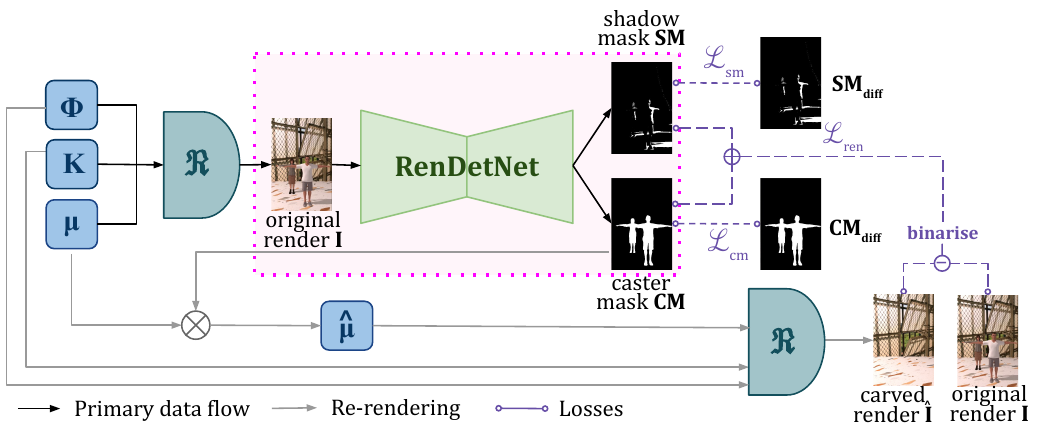}
    \caption{ At train time, we take in lighting $\phi$, camera $\kappa$ and mesh $\mu$ parameters and render the scene $ \textbf{I} =  \mathcal{R}\left(\phi, \kappa, \mu\right)$. We push the render \textbf{I} through our \ours\ to obtain the shadow mask \textbf{SM} and the caster mask \textbf{CM}. We then use \textbf{CM} to carve the mesh and re-render the scene, $ \hat{\textbf{I}} =  \mathcal{R}\left(\phi, \kappa, \hat{\mu}\right)$. \underline{At test time, only the region inside the dotted pink box is used.} }
    \label{our-model}
\end{figure*}

In the following sections, we discuss the \ours\ model and our approach to weakly-supervised training. All elements of the proposed system are shown in Fig.\ \ref{our-model}. At inference, only the parts in the dotted box are required.

\subsection{Differentiable shadow caster verification with \ours}
The model proposed in this paper is a fully-convolutional network based on \cite{wang2018pix2pixHD}. Our \ours\ has a shared encoder $E$ and 2 decoder heads -- one for estimating a shadow mask $D_{sm}$ and one for a caster mask $D_{cm}$. Given a scene render $\textbf{I}$, the process of obtaining a shadow mask $\textbf{SM}$ and a caster mask $\textbf{CM}$ can be formalised as 
\begin{equation}
   \textbf{SM} =  D_{sm}\left(E\left(\textbf{I}\right)\right) \quad\mathrm{and}\quad  \textbf{CM} = D_{cm}\left(E\left(\textbf{I}\right)\right).
    \label{masks}
\end{equation}

The scene is rendered using the rendering function $\mathcal{R}$ that takes lighting $\phi$, camera $\kappa$ and mesh $\mu$ data as arguments, \ie\ \begin{math}
    \textbf{I} =  \mathcal{R}\left(\phi, \kappa, \mu\right).
    \label{rendering-function}
\end{math}
The mesh data can comprise a number of disconnected objects, not necessarily just one. The mesh $\mu$ can be defined as a collection of N triangular faces \{$\mathcal{F}$\} each of which comprises a triplet of 3D vertices ($\textbf{v} \in \mathbb{R}^3$), \ie\
\begin{equation}
    \mu = \left\{
        \mathcal{F}_i 
        \left| 
            \mathcal{F}_i = \{{\textbf{v}_j}\}_{j=1}^3  
        \right. 
    \right\}_{i=1}^N .
    \label{mesh}
\end{equation}

To render the mesh, we need to project the 3D mesh vertices to 2D. To this end, we use a projection function $\pi$ such that 2D point ${\textbf{v}'}$ = \begin{math}
    \pi\left(\textbf{v}, \textbf{K}_{\kappa},\textbf{T}\right) 
\end{math} where \textbf{v}, $\textbf{K}_{\kappa}$ and $\textbf{T}$ signify the 3D point coordinates, the intrinsics of camera $\kappa$ and camera extrinsics respectively. Therefore, the 2D-projected mesh can be described as
\begin{math}
    \mu' =  \left\{
        \mathcal{F}'_i
    \right\}_{i=1}^N
\end{math}
with 2D faces 
\begin{equation}
    \mathcal{F}'_i = \left\{
        \pi(\textbf{v}, \textbf{K}_\kappa,\textbf{T})
        \left|
            \textbf{v} \in \mathcal{F}_i
        \right.
    \right\}
    .
\end{equation}

Next, for our shadow caster verification, we need to check which vertices of the foreground object(s) lie in the estimated \textbf{CM}. To do this, we pick all mesh faces that have at least 1 vertex in this mask. 
More formally, the parts of the mesh inside \textbf{CM} are defined as
\begin{equation}
    \mu'_{cm} = \left\{
        \left.
            \mathcal{F}'_i 
        \right|
        \left(
            \exists \textbf{v} \in \mathcal{F}_i :
                \pi(\textbf{v}, \textbf{K}_\kappa,\textbf{T}) \in \textbf{CM}
        \right)
    \right\}.
\end{equation} 
Having identified the mesh faces identified by the caster mask, we remove them to obtain a new, carved mesh 
\begin{equation}
\hat{\mu} = \mu \backslash \{\mathcal{F}_i | \mathcal{F}'_i \in \mu'_{cm}\}.
\end{equation}
Finally, we update the scene parameters with the new mesh data and re-render the scene. The resulting image \begin{math}
    \hat{\textbf{I}} = \mathcal{R}\left(\phi, \kappa, \hat{\mu}\right)  
\end{math} depicts the scene with some foreground structures masked and correspondingly reduced shadows. 

\subsection{The \ours\ approach to supervision}
To learn to estimate the shadow mask \textbf{SM} and the caster mask \textbf{CM} correctly, we use a number of losses described below. The supervisory signals used for these errors can be obtained at train-time in a self-supervised manner.

The key idea proposed in this paper is the shadow caster verification, \ie\ ensuring that shadows are only detected in the image if they have plausible sources. To achieve this, we calculate the absolute difference between the original render $\textbf{I}$ and the carved mesh render $\hat{\textbf{I}}$. We then apply Otsu thresholding \cite{otsu_1979_threshold} to calculate the threshold $\tau$ and turn this difference into a binary mask, represented by the Iverson brackets $\llbracket{\cdot}\rrbracket$. The change area $\Delta{\textbf{I}}$ stemming from the mesh update can be defined as
\begin{math}
    \Delta{\textbf{I}} =  \llbracket{abs\left(\textbf{I}-\hat{\textbf{I}}\right) > \tau}\rrbracket.
    \label{diff}
\end{math}
We can then use $\Delta{\textbf{I}}$ to calculate the rendering loss $\mathcal{L}_{ren}$ given as
\begin{equation}
    \mathcal{L}_{ren} =  \lVert \Delta{\textbf{I}} - (\textbf{SM} \cup \textbf{CM}) \rVert_1,
    \label{main_loss}
\end{equation}
and use this signal to guide shadow and caster mask generation. 

If $\Delta{\textbf{I}}$ and the masks differ, there are 2 possible explanations: Either a) the portion of the scene we wished to carve was not a shadow-casting foreground object and, thus, there was no $\Delta{\textbf{I}}$ change in this region, or b) the carved caster removed a previously undetected shadow. 
Notably, as $\Delta{\textbf{I}}$ is compared against the sum of the masks, $\mathcal{L}_{ren}$ cannot distinguish between shadows and casters. In other words, $\mathcal{L}_{ren}$ can be trivially satisfied by placing both shadows and casters in \textbf{SM} and leaving \textbf{CM} empty, or vice versa. To better control the masks, we add two extra losses.

Firstly, we add an additional constraint on \textbf{CM} by observing the scene under new lighting $\phi'$ -- constant emitter lighting. The resulting render is equivalent to the scene's reflectance, \ie\ its illumination-invariant characteristics, and objects viewed under such illumination will not cast any shadows or have any self-shadows. If we threshold such a render 
with $\tau'$, we can very easily obtain the object mask, 
\begin{equation}
    \textbf{CM}_{diff} =  \llbracket{
    \mathcal{R}\left(\phi', \kappa, \mu\right)
    > \tau'}\rrbracket,
    \label{cm_gt}
\end{equation}
which can help to ensure that the estimated \textbf{CM} is fully contained within the object region. We can then use this constraint to calculate the caster loss $\mathcal{L}_{cm}$ given as
\begin{equation}
    \mathcal{L}_{cm} = BCE\left(\textbf{CM}_{diff}, \textbf{CM}\right),
    \label{cm_loss}
\end{equation}
where \textit{BCE} denotes the binary cross-entropy loss.

We also restrict the \textbf{SM} generation process in a similar, self-supervised manner; this consists of two steps. Firstly, we consider the scene under 2 different lighting conditions, with a point light flipped along the x-axis (denoted as $-\phi$). 
As before, $\textbf{I}$ denotes the render with original illumination $\phi$ and now $\textbf{I}'$ represents the image with the new light setting, \ie\ $\textbf{I}' = \mathcal{R}\left(-\phi, \kappa, \mu\right)$. 
Under these conditions, the shadow area in \textbf{I} will have a corresponding lighter, non-shadowed region in $\textbf{I}'$ (and vice versa). If we subtract \textbf{I} from $\textbf{I}'$, we will get a positive difference in the \textbf{I} shadow area and a negative difference in the $\textbf{I}'$ shadow area. Since we are only interested in the former, we only consider at the positive part of the difference. We turn the described difference into a binary mask $\textbf{SM}_{diff_1}$ defined as
\begin{equation}
    \textbf{SM}_{diff_1} =  \llbracket{\left(\mathcal{R}\left(-\phi, \kappa, \mu\right)-\textbf{I}\right) > 0}\rrbracket.
    \label{sm_gt1}
\end{equation}
Unfortunately, this way of supervising the \textbf{SM} only works if there is no overlap between the shadows cast under $\phi$ and $-\phi$. To remedy this, we add a second constraint on this self-supervision method. We render our scene with the original mesh $\mu$ and again with just the background (\ie\ with an empty foreground mesh $\mathcal{R}\left(\phi, \kappa, \emptyset\right)$). If we consider the difference between this background render and the original, the region of change will correspond to the objects represented by the mesh $\mu$ and their shadows. Knowing the caster mask estimate $\textbf{CM}_{diff}$ (Eq.\ \ref{cm_gt}), we can obtain the binary mask of just the shadow region,
\begin{equation}
    \textbf{SM}_{diff_2} = \llbracket{abs(
    \mathcal{R}\left(\phi, \kappa, \emptyset\right)-\textbf{I})
    }\rrbracket - \textbf{CM}_{diff} .
    \label{sm_gt2}
\end{equation}
Unlike the previous $\textbf{SM}_{diff_1}$ estimation method, the $\textbf{SM}_{diff_2}$ approach will not show self-cast shadows (as they lie inside $\textbf{CM}_{diff}$). However, this method addresses the shadow overlap issue of the former solution. Therefore, we can combine both techniques and get the full shadow supervision signal defined as
\begin{equation}
    \begin{aligned}
        \textbf{SM}_{diff} = \left(\llbracket{abs(
        \mathcal{R}\left(\phi, \kappa, \emptyset\right)
        -\textbf{I})}\rrbracket - \textbf{CM}_{diff} \right) + 
        \llbracket{\left(\mathcal{R}\left(-\phi, \kappa, \mu\right)-\textbf{I}\right) > 0}\rrbracket.
        \label{sm_gt}
        \end{aligned}
\end{equation}

Given $\textbf{SM}_{diff}$, we can constrain the shadow mask generated by our model. We perform this using the shadow mask loss $\mathcal{L}_{sm}$ defined as
\begin{equation}
    \mathcal{L}_{sm} =  BCE\left(\textbf{SM}_{diff}, \textbf{SM}\right).
    \label{sm_loss}
\end{equation}

The total loss used to guide the training of \ours\ can be described as
\begin{equation}
    \mathcal{L}_{total} = \mathcal{L}_{ren}\lambda_{ren} + \mathcal{L}_{cm}\lambda_{cm} + \mathcal{L}_{sm} \lambda_{sm}.
    \label{total_loss}
\end{equation}
where the $\lambda$s represent the weightings on all loss components. Empirically, we set $\lambda_{cm} = \lambda_{sm} = 1$; $\lambda_{ren}$ is initially set to 0 and then its value increases to 1 after 1000 iterations. 

To evaluate the performance of our model, only the \ours\ part of the described system (pink box in Fig.\ \ref{our-model}) is needed. This also means that at test time no 3D data or other scene parameters are required -- the inference is performed solely on 2D image data.
\section{Experiments}
In the following sections, we discuss the datasets created as part of this publication and use them to evaluate the performance of a number of models.

\subsection{The dataset}
As previously discussed, capturing training datasets for shadow detection is a challenge. Recently, Inoue \& Yamasaki \cite{inoue2020learning} proposed a pipeline for synthetic shadow removal/detection dataset creation. In their setup, shadow masks are used to cast shadows of varying intensity onto previously shadow-free scenes, thus generating shadowed samples. While interesting, this approach does not help to resolve the `dark vs shadow' issue faced by the existing solutions. 

To address this, we set out a few objectives for our data generation pipeline: We want to include information about shadows and their casters. We do not want to rely on any existing shadow masks (or shadow-filtered data) to run the generation pipeline. Finally, we want to create the dataset in a fully automated, self-supervised manner, without any human annotation. Hand-labelling datasets limits their scalability, so learning with imperfect labels should be prioritised. A number of learning-based solutions already get their (imperfect) masks from previously trained models, so we consider our physics-backed estimation approach to be similarly viable while also being capable of adapting to new data.

To develop and demonstrate our idea, we rendered 2 datasets using \textit{Mitsuba 3} \cite{Mitsuba3}; we refer to them as Datasets \#1 \& \#2. Due to copyright reasons, the meshes used to create Dataset \#1 cannot be released. In the interest of reproducibility, we created Dataset \#2 using only publicly available components and we provide the code used to create it. 

Each dataset scene features between 1 and 3 textured meshes -- human meshes from 3D Virtual Humans (3DVH) \cite{caliskan2020multi} for Dataset \#1 or a mix of BEDLAM \cite{bedlam} humans and clutter objects \cite{dataset-object2,dataset-objects1} for Dataset \#2. Dataset \#1 features humans in fixed poses. To maximise the cast shadow diversity, the humans in Dataset \#2 are posed in a variety of ways using AGORA \cite{patel2021agora}. The objects (human or not) are placed in a range of locations on patterned floors and in front of heavily textured backgrounds, the textures of which come from \cite{tiles} and \cite{bkgd_20k,zhou_2017_places} respectively. We additionally vary the scene illumination, changing both the magnitude of the constant emitter as well as the positioning and brightness of a point light. The resulting datasets feature approx.\ 4.2k train / 900 test samples (Dataset \#1) and 9k train / 1k test samples (Dataset \#2).

\subsection{Evaluation metrics and experimental setup}
The \ours\ system was developed in Pytorch. The first version, with Dataset \#1, was trained for approx.\ 8k steps on a GeForce RTX 3090. More compact and diverse Dataset \#2 required approx.\ 12k steps on a single GeForce RTX 2080. All variants of \ours\ were trained in the Lab colour space unless otherwise stated. The evaluations in the following sections are performed in terms of the BER (Balanced Error Rate) score, defined as
\begin{equation}
    BER = \left(1 - 0.5 \times \left( \frac{TP}{N_s} + \frac{TN}{N_{ns}} \right)\right) \times 100,
    \label{ber}
\end{equation}
where TP/TN symbolise the true positives/negatives and $N_s$/$N_{ns}$ describe the total number of shadow/non-shadow pixels. We also cite the average shadow region (S) and non-shadow region (NS) BER scores. In all cases, a lower BER score signifies a better result. When quantitative data is reported, the best score is shown in \textbf{bold} and the runner-up is \underline{underlined}. All training and evaluation runs were performed on full-size images, \ie\ $512 \times 768$ pixels.

\subsection{Ablation study}
First, we conduct an ablation study to demonstrate the impact of each of our design choices. Specifically, we look at the impact of different training losses and discuss the importance of choosing the right colour space for training a system for lighting-related problems. At the end, we also demonstrate the shadow-caster matching performance of RenDetNet.

\textbf{Model losses study.}
In this study, we gradually add losses to our model trained on Dataset \#1 and measure their impact on \ours's performance. The quantitative results of this evaluation are shown in Table \ref{loss_ablation}. During the experiments, we first train the most basic version of our model. As the rendering loss $\mathcal{L}_{ren}$ does not provide enough guidance to disentangle \textbf{SM} and \textbf{CM} on its own, we consider our ``bare minimum'' scenario to operate based on $\mathcal{L}_{ren}$ and $\mathcal{L}_{sm}$ (line \#1). We then add $\mathcal{L}_{cm}$ (line \#2) to control both estimated masks, which leads to slight improvements, particularly in the shadowed region. Next, we consider the network trained with just $\mathcal{L}_{cm}$ and $\mathcal{L}_{sm}$ (line \#3) and no shadow caster verification. The resulting model is good at spotting shadows (lowest BER(S)) but it struggles in the non-shadow regions (highest BER(NS)). We attribute this to the dark/textured regions that are misclassified as shadows. Finally, to combine the success of models \#2 and \#3, we adopt a staged training mechanism: We start training the model with just the mask losses and add $\mathcal{L}_{ren}$ after 1000 iterations. Around this time the \textbf{SM} and \textbf{CM} learning process slows down, but the masks have become plausible enough to be constrained in a physics-backed manner. The resulting system (line \#4) has the lowest overall BER score, which confirms the complementary nature of the previously discussed losses.

\begin{table*}[h]
    \caption{Loss ablation study - Dataset \#1.}
    \begin{center}
        \begin{tabular}{ cccccccc  }
            \hline
            \# & $\mathcal{L}_{ren}$  & $\mathcal{L}_{sm}$ & $\mathcal{L}_{cm}$ & staged &  BER $\downarrow$ & BER (S) $\downarrow$& BER (NS) $\downarrow$\\
            \hline
            1&\checkmark & \checkmark & & & 7.324 & 13.526 & \textbf{1.123} \\
            2&\checkmark &\checkmark & \checkmark&  & 7.182 & 13.103 & \underline{1.261} \\
            3&  & \checkmark &\checkmark & &\underline{6.598} & \textbf{8.939} & 4.257 \\
            4&\checkmark & \checkmark& \checkmark& \checkmark& \textbf{6.344} & \underline{11.010} & 1.679 \\
            \hline
        \end{tabular}
    \end{center}
    \label{loss_ablation}
\end{table*}

\textbf{Colour space study.}
Most shadow detection models are trained in the RGB colour space. However, since shadows are lighting-related phenomena, we experimented with training our model in the \textit{Lab} colour space. In \textit{Lab}, the \textit{L} channel corresponds to illumination and the colour information is captured by the \textit{a} and \textit{b} channels. As demonstrated in Table \ref{Lab_ablation}, training \ours\ on data with such a representation is clearly advantageous.

\begin{table}[h]
    \caption{Colour space ablation study - Dataset \#1.}
    \begin{center}
        \begin{tabular}{ cccc  }
            \hline
            Colour space &  BER $\downarrow$ & BER (S) $\downarrow$ & BER (NS) $\downarrow$ \\
            \hline
            RGB & \underline{46.823} & \underline{87.269} & \underline{6.377}\\
            Lab & \textbf{6.344} & \textbf{11.010} &\textbf{1.679} \\
            \hline
        \end{tabular}
    \end{center}
    \label{Lab_ablation}
\end{table}

\begin{figure}[b]
    \begin{center}
    \small
        \begin{tabular}{  c @{\hspace{0.05cm}} c @{\hspace{0.05cm}} c @{\hspace{0.3cm}}c @{\hspace{0.05cm}} c @{\hspace{0.05cm}} c }
        Input& Estim.\ CM &  Estim.\ SM & Input& Estim.\ CM &  Estim.\ SM\\
        \includegraphics[width=1.9cm]{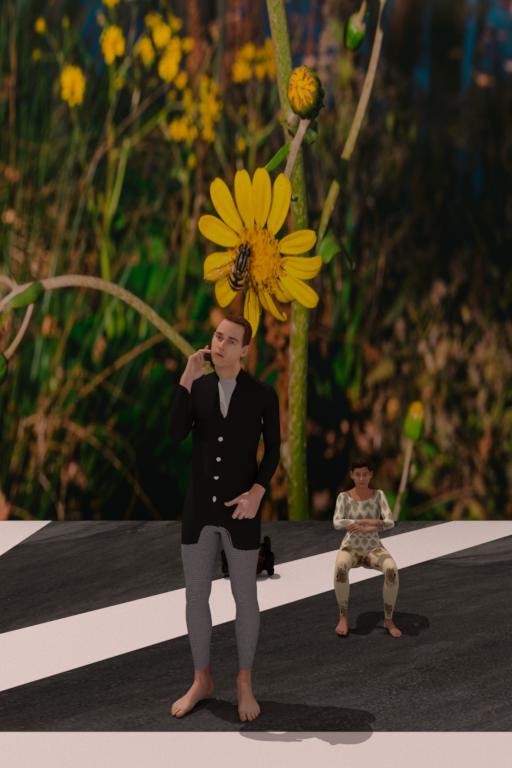}&
        \includegraphics[width=1.9cm]{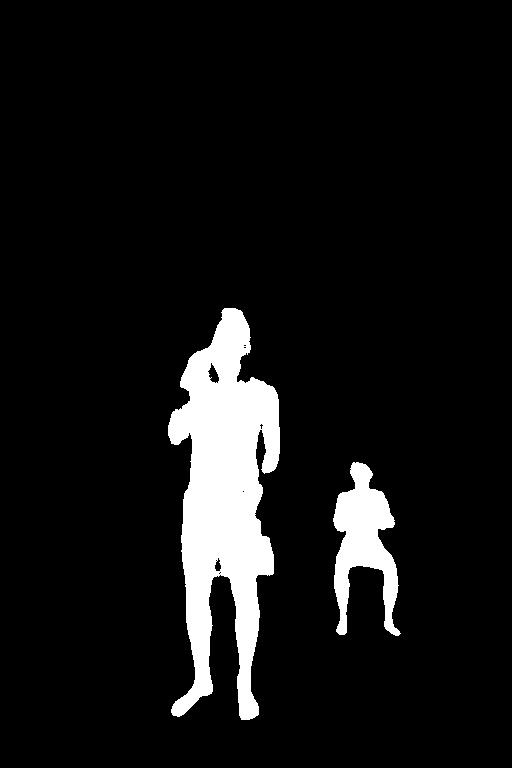}&
        \includegraphics[width=1.9cm]{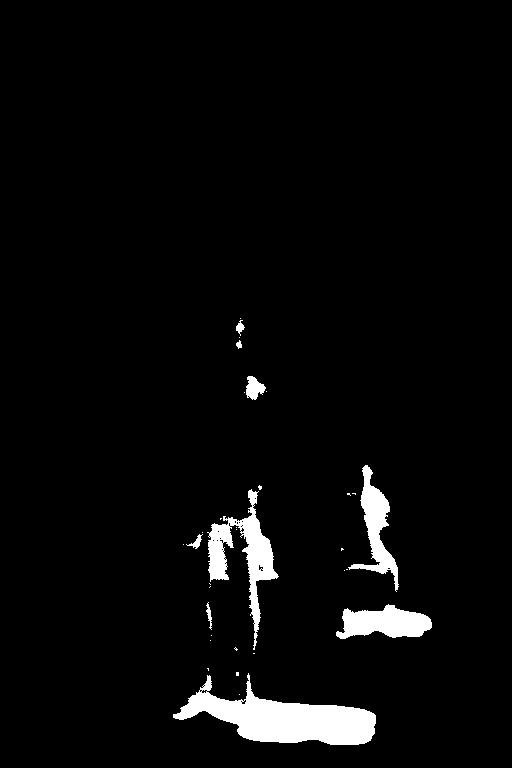}&
        
        \includegraphics[width=1.9cm]{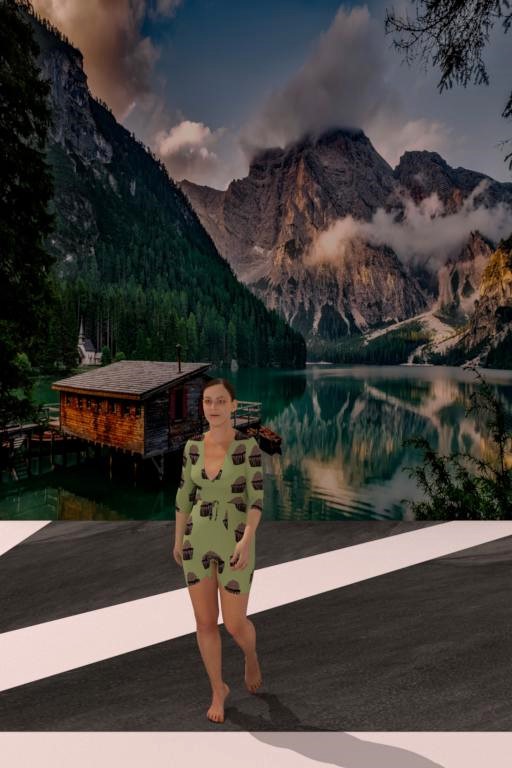}&
        \includegraphics[width=1.9cm]{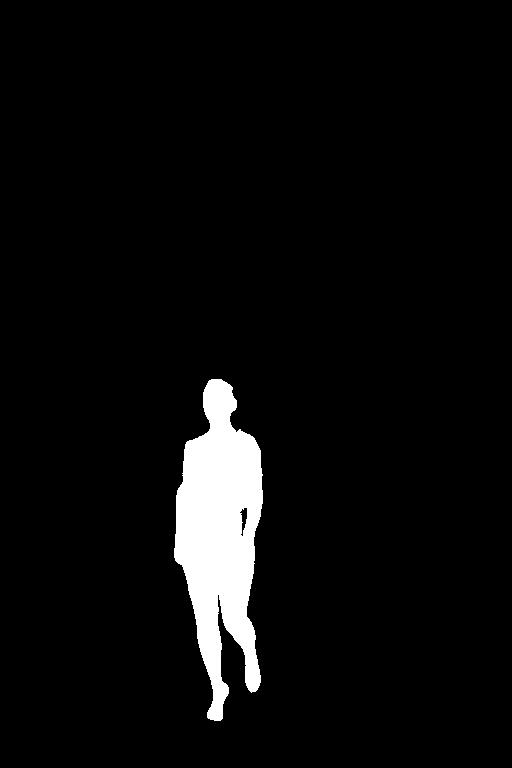}&
        \includegraphics[width=1.9cm]{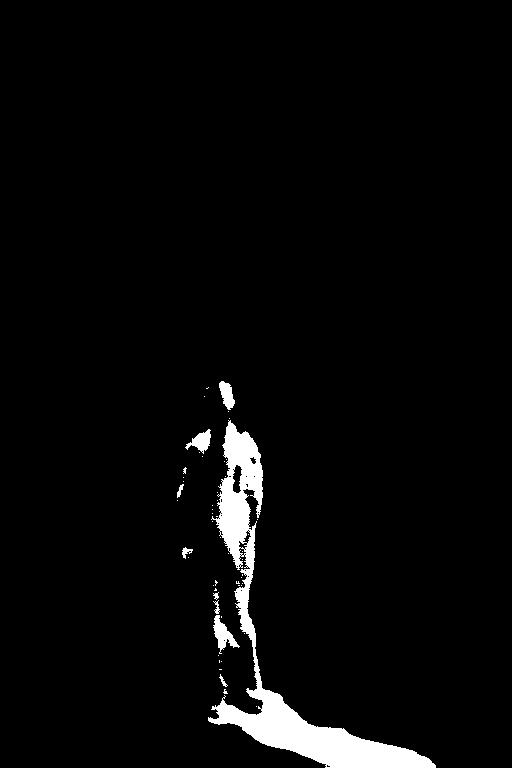}\\
        \end{tabular}
    \end{center}
\caption{Examples of RenDetNet's caster \& shadow identification abilities - Dataset \#2}
\label{caster}
\end{figure}

\textbf{Shadow-caster pairing. } As demonstrated in Fig.\ \ref{caster}, aside from detecting shadows, the proposed RenDetNet is also capable of identifying the shadows' casters within the scene with high certainty and a good amount of detail. 

\subsection{Comparisons with existing models}
\begin{figure*}[h]
    \begin{center}
    \scriptsize
        \begin{tabular}{ c @{\hspace{0.05cm}} c @{\hspace{0.05cm}} c @{\hspace{0.05cm}} c @{\hspace{0.05cm}} c @{\hspace{0.05cm}} c }
        Input & Ground Truth & DSC \cite{hu2018direction} & BDRAR \cite{zhu2018bidirectional}& FDRNet \cite{zhu2021mitigating} & \ours\ (ours) \\
    
        \includegraphics[width=1.9cm]{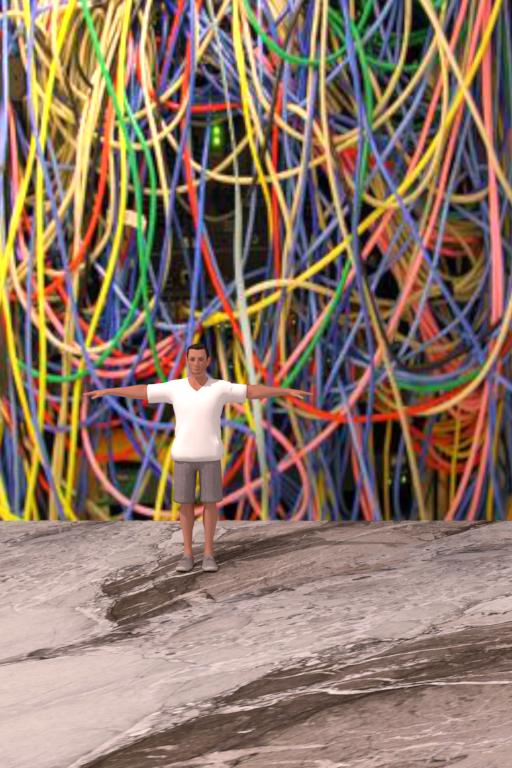}&
        \includegraphics[width=1.9cm]{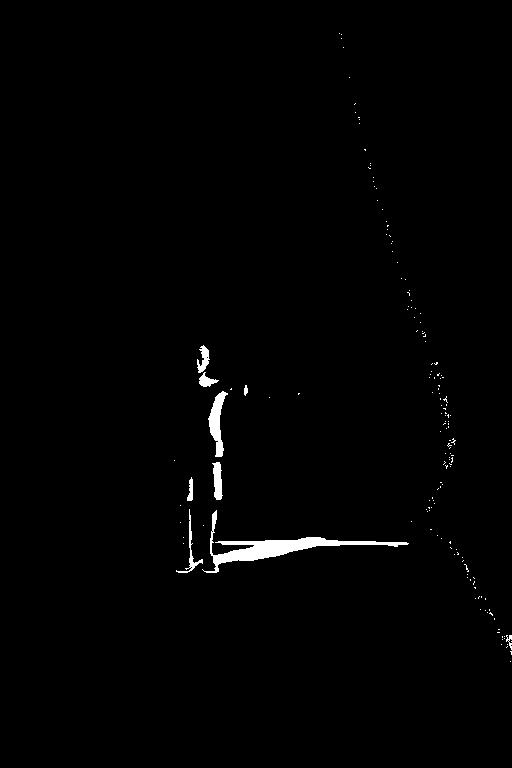}&
        \includegraphics[width=1.9cm]{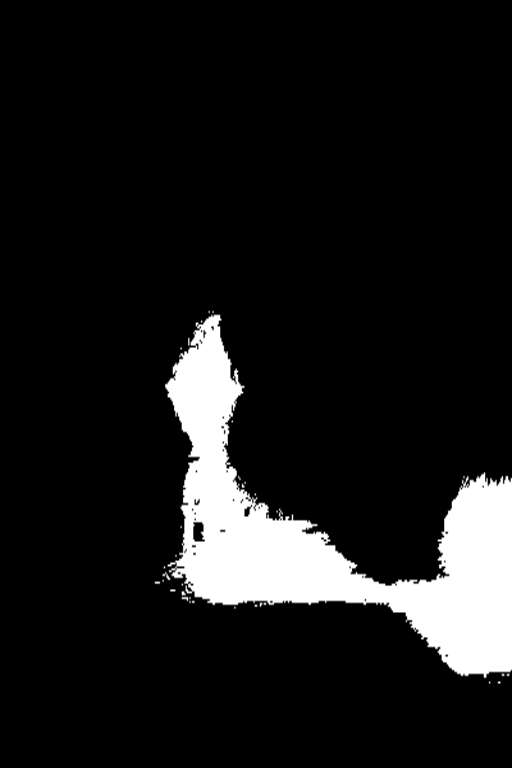}&
        \includegraphics[width=1.9cm]{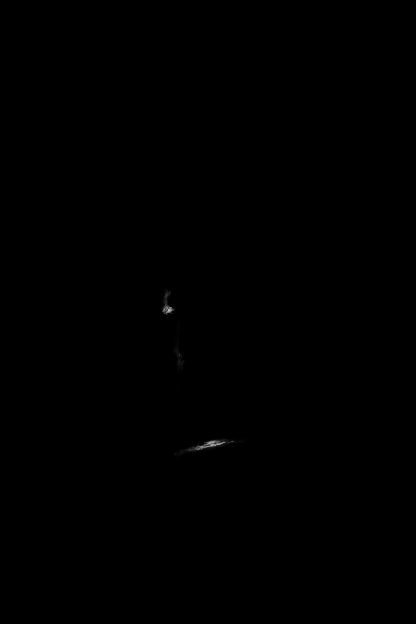}&
        \includegraphics[width=1.9cm]{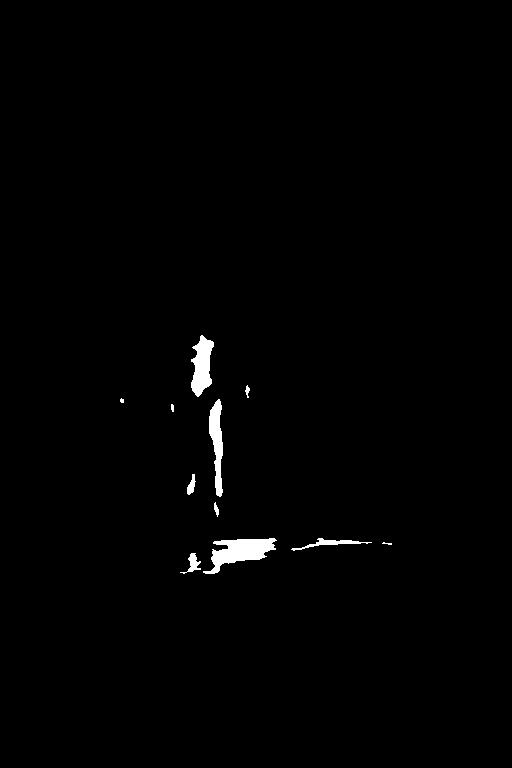}&
        \includegraphics[width=1.9cm]{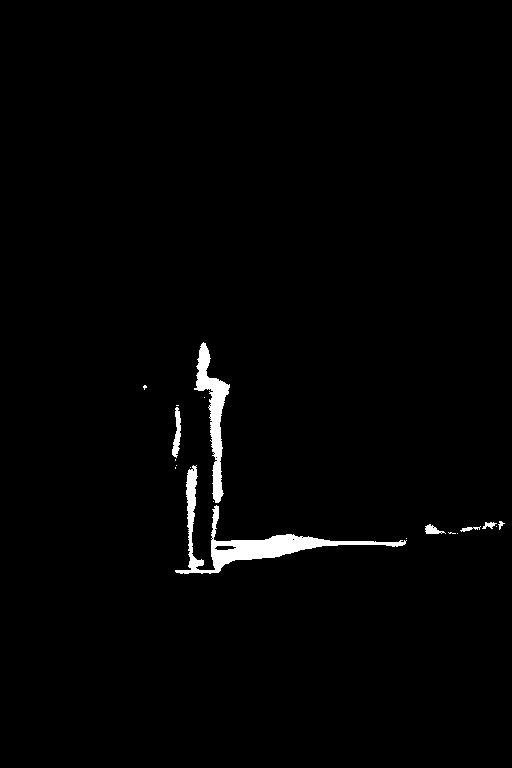}\\
    
        \includegraphics[width=1.9cm]{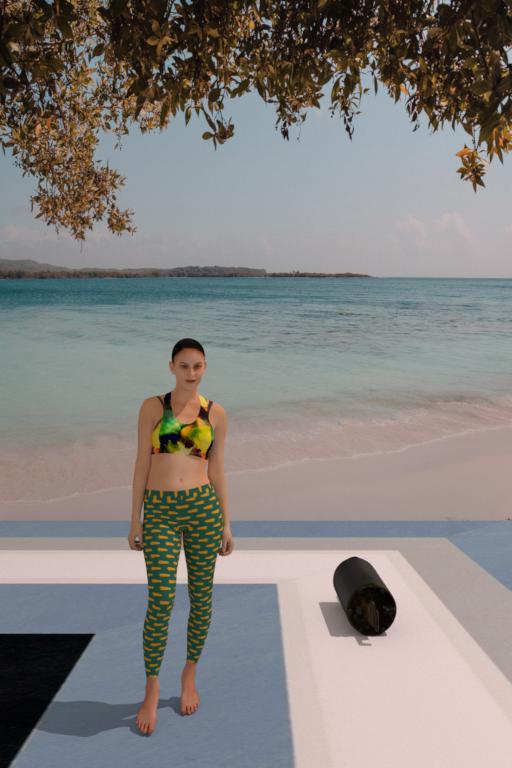}&
        \includegraphics[width=1.9cm]{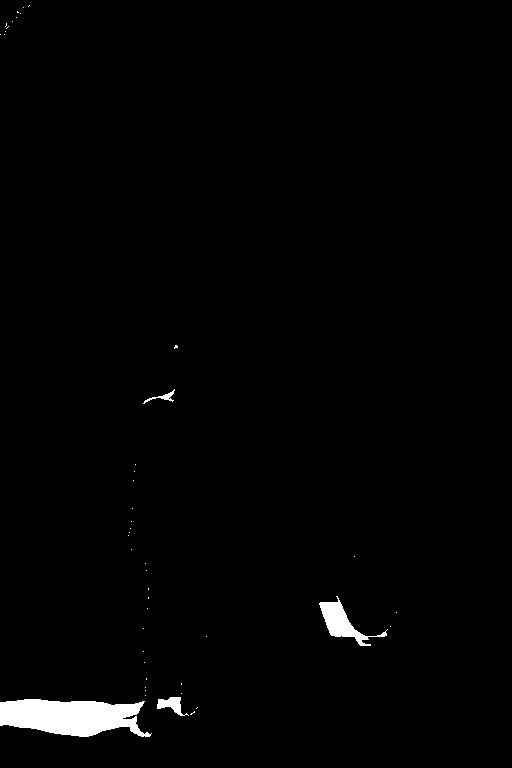}&
        \includegraphics[width=1.9cm]{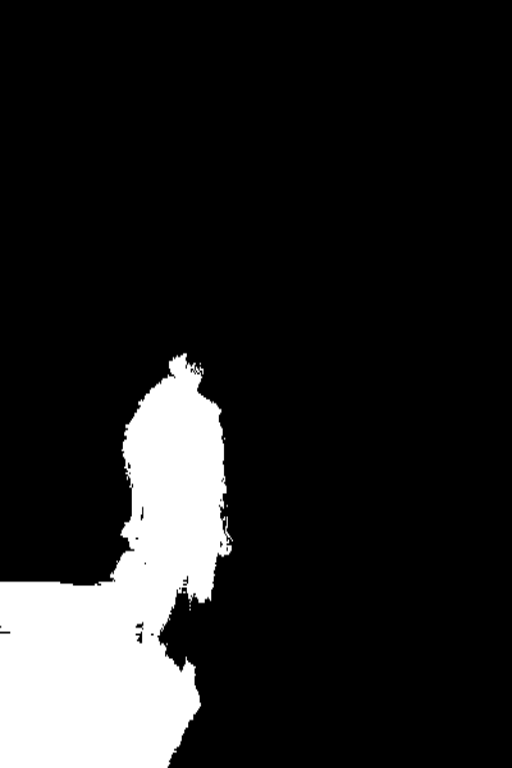}&
        \includegraphics[width=1.9cm]{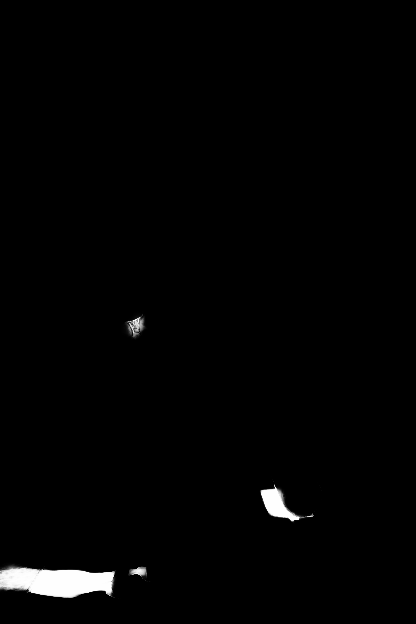}&
        \includegraphics[width=1.9cm]{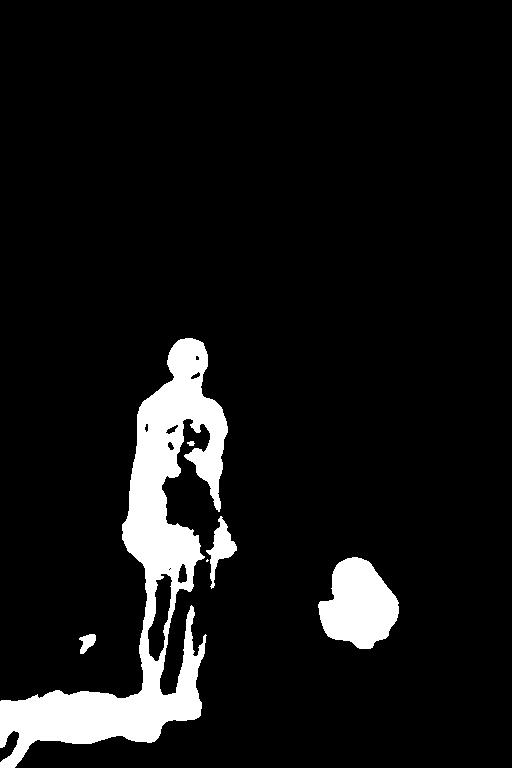}&
        \includegraphics[width=1.9cm]{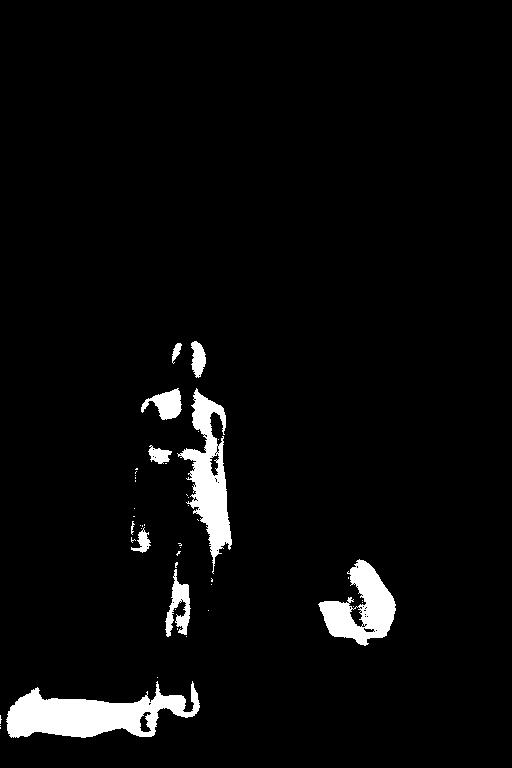}\\
    
        \includegraphics[width=1.9cm]{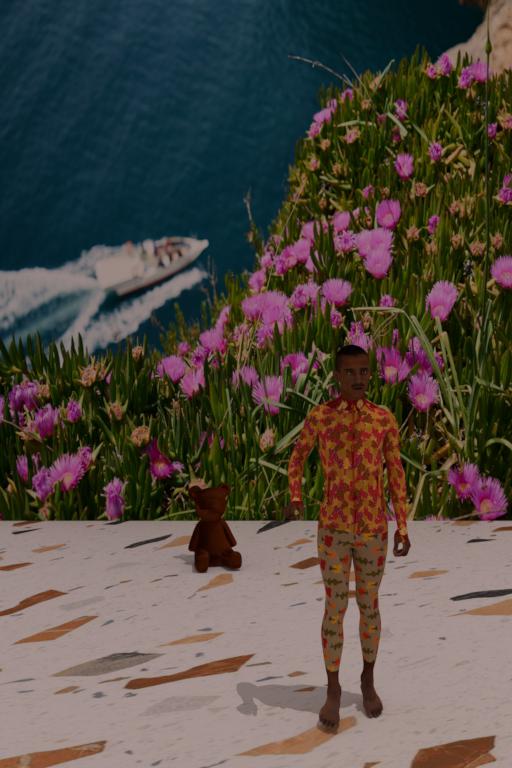}&
        \includegraphics[width=1.9cm]{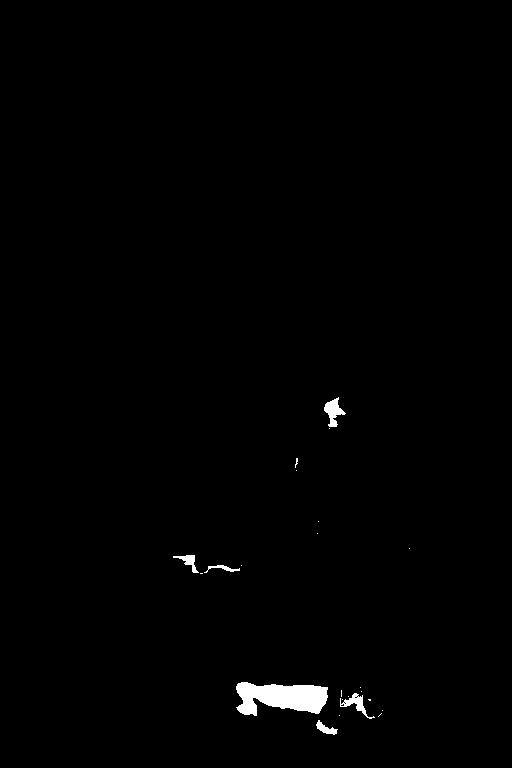}&
        \includegraphics[width=1.9cm]{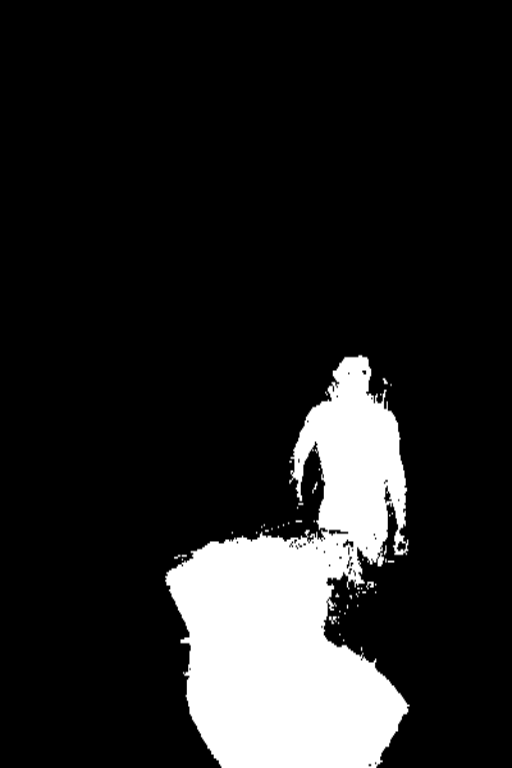}&
        \includegraphics[width=1.9cm]{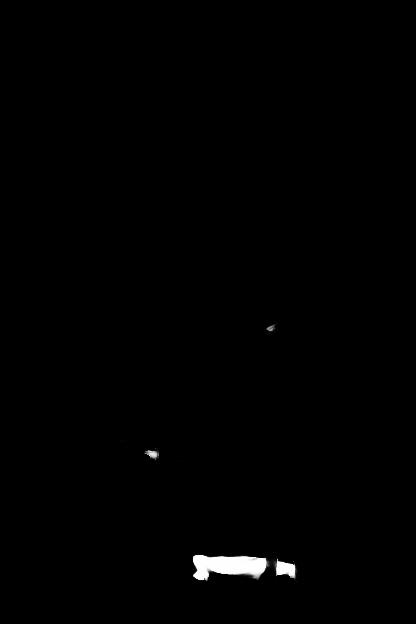}&
        \includegraphics[width=1.9cm]{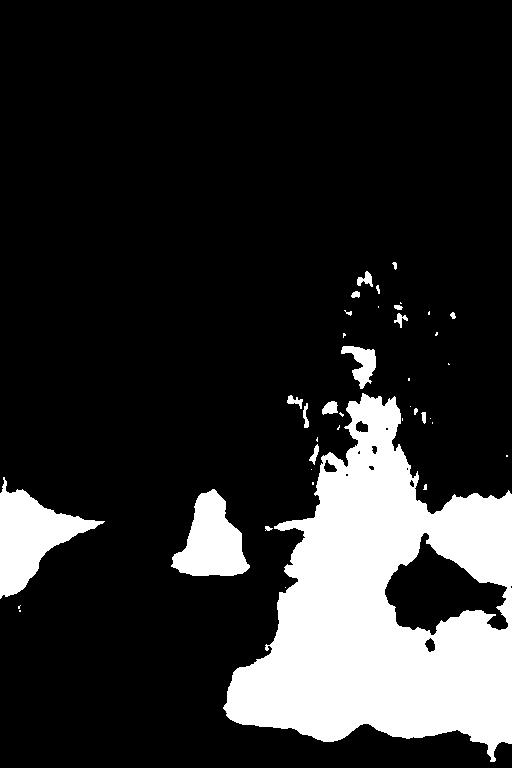}&
        \includegraphics[width=1.9cm]{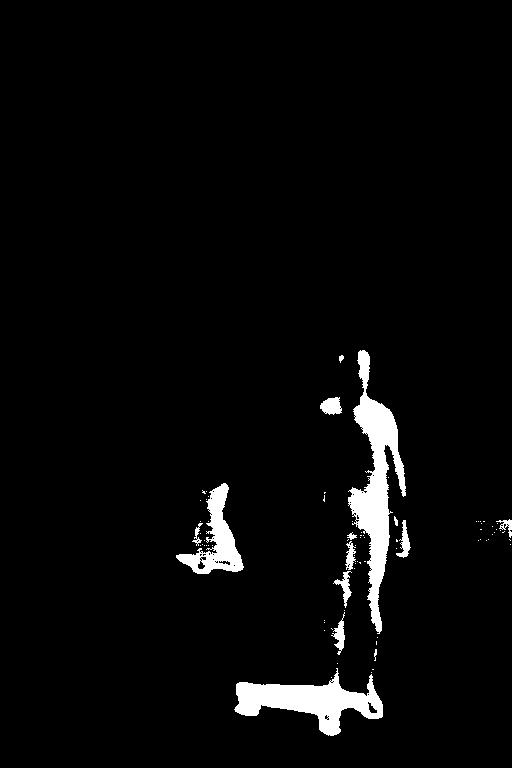}\\
    
        \end{tabular}
    \end{center}
\caption{Qualitative results -- row 1: Dataset \#1; rows 2-3: Dataset \#2. }
\label{visual_our_data}
\end{figure*}

For fair comparison with state-of-the-art solutions, we rendered all of our training images and generated a shadow mask for each sample to create supervised 2D versions of our datasets. Some of the recent shadow detection models rely on false-negatives and false-positives pre-calculated based on the outputs of existing models, \eg\ \cite{zheng2019distraction}, or require more data for different multi-task problems \cite{chen2020multi} or sub-problems \cite{wang2,wang_2022_inst_tpami,wang1}, which unfortunately makes them infeasible for our evaluation. Therefore, for our comparisons, the compatible models -- DSC \cite{hu2018direction}, BDRAR \cite{zhu2018bidirectional} and FDRNet \cite{zhu2021mitigating} -- were trained on our data using their publicly available code. The results of this study are presented in Table \ref{sota_our_data} and Fig.\ \ref{visual_our_data}.

The qualitative data shows the advantage of our caster verification step. In the top row of Fig.\ \ref{visual_our_data} (Dataset \#1) we can see that \ours\ manages to find the shadows, both cast on the ground as well as on the caster itself. FDRNet performs similarly well yet loses some detail in the shadow cast on the floor. BDRAR struggles to find any shadows while DSC finds the caster but also estimates a large portion of the floor as shadow. We speculate this is due to the clutter and variety of colours and textures in the scene. In the next two rows (Dataset \#2), DSC largely confuses the caster and the shadow, loosely covering both areas. BDRAR correctly identifies the cast shadows, but completely ignores any self-cast shadows. FDRNet captures more details and self-cast shadows than BDRAR, yet, similarly to DSC, its shadow boundaries tend to spill. Our model reconstructs the ground shadows to a similar degree as BDRAR, and preserves more detail than FDRNet and DSC.

\begin{table}[b!]
    \caption{Performance in terms of BER - Dataset \#1 (top) \& \#2 (bottom) }
    \begin{center}
        \begin{tabular}{ ccccc }
            \hline
            Dataset&Model & BER $\downarrow$ & BER (S) $\downarrow$ & BER (NS) $\downarrow$ \\
            \hline
            \#1 &    DSC \cite{hu2018direction}& 14.716 & 19.256 & 10.176\\
            &BDRAR \cite{zhu2018bidirectional} & 8.619 & 15.887 & \textbf{1.352} \\
            &FDRNet \cite{zhu2021mitigating} & \underline{7.276}  & \textbf{10.714} & 3.839\\
            &\ours\ (ours) &\textbf{6.344} & \underline{11.010} &\underline{1.679} \\
            
            \hline 
            \#2 &DSC \cite{hu2018direction}& 15.029 & \underline{16.599} & 13.459\\
            &BDRAR \cite{zhu2018bidirectional} & 15.415 & 30.359 & \textbf{0.472} \\
            &FDRNet \cite{zhu2021mitigating} & \textbf{11.634}  & \textbf{14.449} & 8.820\\
            &\ours\ (ours) & \underline{11.992} & 21.347 & \underline{2.638} \\
            \hline
        \end{tabular}
    \end{center}
    \label{sota_our_data}
\end{table}

Numerically, our \ours\ achieves a good balance in terms of BER(S) vs BER(NS). This places it as the top performer (Dataset \#1) or a close runner-up (Dataset \#2); nevertheless, visually, our method clearly outperforms the top-performing FDRNet. Conservative in its shadow estimation, BDRAR tends to be superior in terms of BER(NS), yet the system tends to underestimate the shadows (high BER(S)). In contrast, FDRNet tends to overestimate the shadows, which then leads to good precision in terms of shadow-area detection, but worse BER(NS) performance. \ours\ offers a happy medium.

\section{Conclusions}
In this paper, we have proposed a new weakly-supervised approach to physically-verified shadow caster matching. The method allows us to accurately estimate shadows, both cast on the external environment as well as self-cast shadows which overlap the caster itself. Additionally, the proposed \ours\ model can be trained in a fully-differentiable setup with the supervisory signals calculated on the fly, and requires no hand annotation or other human input. Finally, we have presented a new self-supervised caster-aware dataset generation pipeline. While developed on a particular set of images, meshes and textures, the same idea could be applied to other sets of data.

As future work, it would be interesting to explore alternative training regimes which can jointly exploit existing strongly supervised datasets alongside our novel self-supervised data generation approach. It would also be productive to obtain a set of real-life images, perhaps with hand-annotated masks, to create an improved validation set for shadow/caster estimation.

\subsection*{Acknowledgements} This work was partially supported by the British Broadcasting Corporation (BBC) and the Engineering and Physical Sciences Research Council's (EPSRC) industrial CASE project ``Computational lighting in video'' (voucher number 19000034).

\clearpage
{\small
\bibliographystyle{splncs04}
\bibliography{main}
}

\end{document}